\documentclass{Interspeech}

\interspeechcameraready

\title{The NTNU System at the S\&I Challenge 2025 SLA Open Track}

\author[affiliation={1}]{Hong-Yun}{Lin}
\author[affiliation={1}]{Tien-Hong}{Lo}
\author[affiliation={2}]{Yu-Hsuan}{Fang}
\author[affiliation={1}]{Jhen-Ke}{Lin}
\author[affiliation={1}]{Chung-Chun}{Wang}
\author[affiliation={1}]{Hao-Chien}{Lu}
\author[affiliation={1}]{Berlin}{Chen}

\affiliation{Department of Computer Science and Information Engineering}{National Taiwan Normal University}{Taiwan}
\affiliation{Institute of AI Interdisciplinary Applied Technology}{National Taiwan Normal University}{Taiwan}
\email{\{buffett, teinhonglo, andyfang, jacob, takala, howchien, berlin\}@ntnu.edu.tw}


\usepackage{xcolor} 
\usepackage{refcount} 

\usepackage{graphicx} 

\usepackage{comment}
\usepackage{subcaption} 
\usepackage{amsmath} 
\newcommand{\good}{\ensuremath{\uparrow}} 
\newcommand{\bad}{\ensuremath{\downarrow}} 

\begin{document}

\maketitle

\begin{abstract}
A recent line of research on spoken language assessment (SLA) employs neural models such as BERT and wav2vec 2.0 (W2V) to evaluate speaking proficiency across linguistic and acoustic modalities. Although both models effectively capture features relevant to oral competence, each exhibits modality-specific limitations. BERT-based methods rely on ASR transcripts, which often fail to capture prosodic and phonetic cues for SLA. In contrast, W2V-based methods excel at modeling acoustic features but appear to lack semantic interpretability. To overcome these limitations, we propose a system that integrates W2V with Phi-4 multimodal large language model (MLLM) through a score fusion strategy. The proposed system achieves a root mean square error (RMSE) of 0.375 on the official test set of the Speak \& Improve Challenge 2025, securing second place in the competition. For comparison, the RMSEs of the top-ranked, third-ranked, and official baseline system are 0.364, 0.384, and 0.444, respectively.
\end{abstract}

\section{Introduction}\label{sec:intro}

With the rapid advancement of computer technology and the increasing global population of second language (L2) learners, spoken language assessment (SLA) has garnered significant attention, particularly within the domain of computer-assisted language learning (CALL). SLA systems are designed to provide timely and informative feedback on learners' oral performance, thereby facilitating autonomous and low-anxiety practice aimed at enhancing spoken language proficiency. Moreover, such systems can reduce the instructional burden on language educators while offering a more objective and consistent evaluation of L2 learners' speaking abilities. In light of recent progress in human language technologies, SLA systems have been widely integrated into CALL environments to support and enhance the language acquisition process.

The development of spoken language assessment (SLA) has undergone several paradigm shifts, each reflecting advancements in representation learning and modeling strategies. Initial efforts primarily relied on hand-crafted acoustic features, often coupled with forced alignment techniques to extract temporal phonetic information. A subsequent paradigmatic shift introduced textual self-supervised embeddings, such as BERT \cite{Devlin2019BERT, WangEQM21}, which enhanced content-level assessment. However, these models are inherently dependent on automatic speech recognition (ASR) outputs, rendering them susceptible to transcription errors and unable to fully capture prosodic or phonetic nuances \cite{Banno2022L2PA, raina20_interspeech}. To overcome these limitations, acoustic self-supervised models like wav2vec 2.0 \cite{Baevski20-Wav2Vec2} emerged, offering the ability to process raw speech directly. When fine-tuned for tasks such as proficiency estimation \cite{BannoM_SLT2022} or pronunciation evaluation \cite{kim22k_interspeech}, these models have demonstrated superior robustness under noisy ASR conditions \cite{banno23_slate}, although they often fall short in capturing higher-level linguistic properties, such as lexical appropriateness or syntactic coherence. Addressing this gap, recent work has introduced multimodal large language models (MLLMs) \cite{Tang2024SALMONN, hu-etal-2024-wavllm}, which integrate speech encoders, modality adapters, and decoder-only language models to enable multi-modal processing. These models jointly consider both the linguistic content and delivery characteristics of spoken input, achieving competitive performance on standard benchmarks \cite{hu-etal-2024-wavllm} and offering a promising direction for unifying the strengths of earlier paradigms. In parallel, automated essay scoring (AES) has seen rapid progress through the application of large language models (LLMs)\cite{mansour-etal-2024-large, shibata2025lceszeroshotautomatedessay}, with GPT-style systems showing strong generalization in zero- and few-shot settings. However, the adoption of MLLM-based approaches for holistic scoring in SLA remains relatively limited and underexplored.

In response to these challenges, the Speak \& Improve Challenge 2025 provides a timely and realistic benchmark for gauging the performance of spoken language assessment systems under constraints of limited data and computational resources \cite{Qian25-SIChallenge}. Our proposed two systems, dubbed NTNU SMIL V and NTNU SMIL X, are designed to address this objective by integrating two distinct graders that operate on the same spoken input to estimate the speaking proficiency of the learner. The first component is a wav2vec 2.0-based model that focuses on acoustic characteristics, including pronunciation and fluency, while the second is a Phi-4-based multimodal large language model (MLLM) that extracts semantic and syntactic information from the speech signal to support accurate proficiency scoring. The predictions generated by these two graders are combined using a score fusion strategy, wherein the fusion weights are selected to minimize root mean square error (RMSE) on the development set. Our best-performing configuration attained a root mean square error (RMSE) of 0.375, ranking second in the competition. For reference, the RMSEs of the first-place, third-place, and official baseline systems were 0.364, 0.384, and 0.444, respectively. This performance highlights the effectiveness of combining a self-supervised speech encoder with an MLLM component for spoken language assessment.

\begin{figure*}[!t]
    \centering
    \subfloat[]{
        \includegraphics[width=0.23\linewidth]{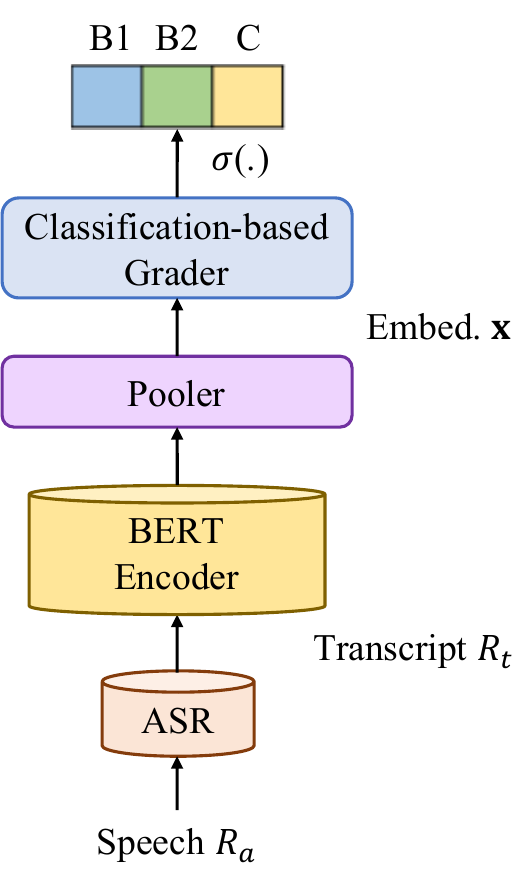}%
    \label{fig:bert_arch}}
    \hfil
    \subfloat[]{
        \includegraphics[width=0.23\linewidth]{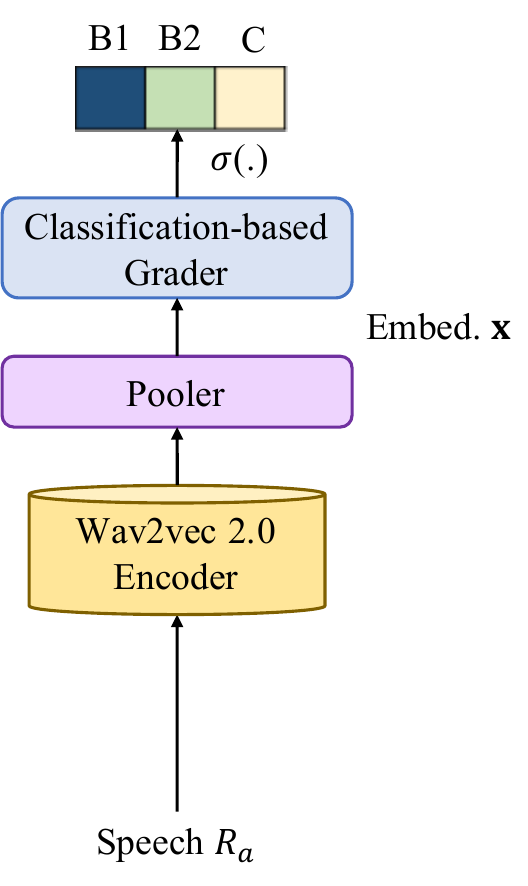}%
    \label{fig:wav2vec_arch}}
    \hfil
    \subfloat[]{
        \includegraphics[width=0.49\linewidth]{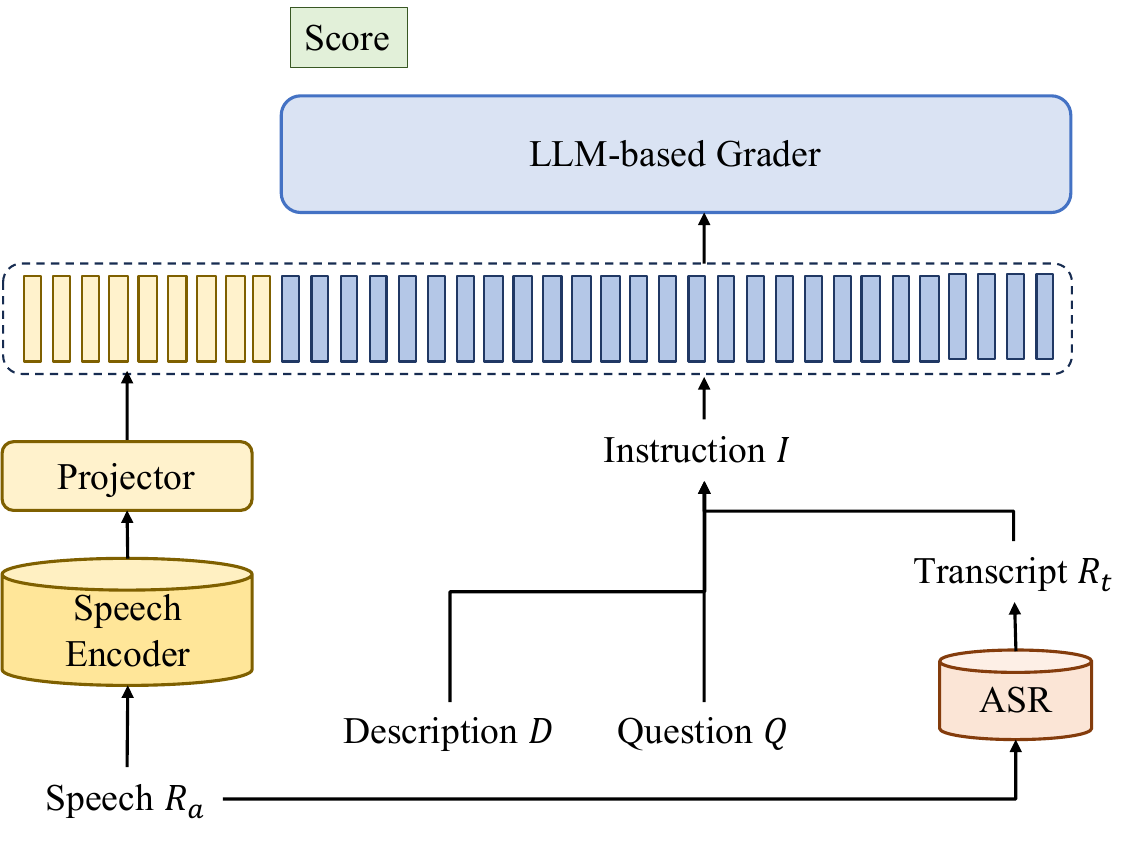}%
    \label{fig:mllm_grader}}
    \caption{Architectures for spoken language assessment (SLA). (a) Text-based grader: speech ($R_a$) is transcribed to text ($R_t$) and processed by a BERT encoder. (b) Speech-based grader: raw speech ($R_a$) is directly encoded using wav2vec 2.0. (c) MLLM-based grader: inputs include speech ($R_a$), ASR transcript ($R_t$), description ($D$), and question ($Q$), which are combined into an instruction ($I$) for the MLLM-based grader.}
\label{fig:text_speech_graders}
\end{figure*}

\section{Method}\label{sec:method}

Spoken language assessment (SLA) has traditionally capitalized on text-based models; iconic ones include those built on BERT, which process transcripts generated by ASR systems. These models are effective at evaluating content-related aspect of speech (i.e., what is said) and have formed the basis of several benchmark systems, including the official challenge baseline \cite{Qian25-SIChallenge}. A typical BERT architecture is shown in Figure \ref{fig:bert_arch}. However, their performance is inevitably limited by ASR errors, which can distort the intended content, and by their inability to capture prosodic and acoustic features critical to speaking proficiency, such as intonation, rhythm, and stress \cite{Devlin2019BERT, Banno2022L2PA}.
Recent work has turned to speech foundation models such as wav2vec 2.0, which process raw audio directly to overcome the aforementioned shortcomings. These models are well-suited to capturing the delivery-related aspect of speech (i.e., how something is said) and are capable of modeling fine-grained acoustic information. A typical wav2vec 2.0 architecture is shown in Figure \ref{fig:wav2vec_arch}. Nevertheless, while strong in acoustic modeling, these systems often lack the capacity to fully interpret semantic content without explicit linguistic grounding~\cite{Banno2022L2PA}. To address the respective limitations of both BERT- and wav2vec 2.0-based methods, we investigate the use of multimodal large language models (MLLMs), which integrate both acoustic and textual inputs. As illustrated in Figure~\ref{fig:mllm_grader}, our proposed MLLM-based grader leverages the semantic reasoning capabilities of large language models alongside rich acoustic features, providing a unified modeling framework for comprehensive and robust SLA.

\subsection{System architecture}

Our work aims to reinforce the comprehensive strengths of SLA on both acoustic and  language understanding of spoken responses. To this end, we first employ a robust wav2vec 2.0-based speech grader to establish a strong foundation in assessing speech proficiency. On the separate front, we introduce an advanced multimodal large language model (MLLM), Phi-4-multimodal, to not only provide a holistic assessment by jointly processing speech and text but also to enhance the system's understanding of linguistic content inherent in spoken responses. This MLLM is designed to jointly process text, image, and audio inputs within a single neural network, enabling a deeper, more aligned multimodal understanding. Finally, a score-conditioned fusion mechanism integrates the outputs of these two specialized graders.

\subsection{Speech grader}

We draw on the pre-trained wav2vec 2.0 (W2V) encoder \cite{Baevski20-Wav2Vec2} to extract frame-level acoustic representations from raw speech input. Specifically, given an utterance corresponding to part $p$, we process the raw waveform $R_{A_p}$ through the W2V model to obtain a sequence of latent vectors $\mathbf{h}_{0:T-1} \in \mathbb{R}^d$, where $T$ denotes the length of the sequence. To derive an utterance-level representation, we apply attention pooling \cite{safari20_interspeech} over the frame-level features:

\begin{align}
\mathbf{x} = \text{AttnPool}(\mathbf{h}_{0:T-1}).
\end{align}

We then incorporate a prototypical embedding module \cite{lo-etal-2024-effective}, in which each learnable prototype vector $\mathbf{p}_j$ represents level $j$ and is initialized with the mean embedding of training examples labeled accordingly. The similarity between the utterance vector $\mathbf{x}$ and each prototype is computed using cosine similarity, resulting in a prototype-informed vector $\mathbf{s} = [s_1, s_2, \dots, s_N] \in \mathbb{R}^N$, where $N$ is the number of CEFR levels \cite{CefrCouncil2001_common} (i.e., $N=8$ in the Speak \& Improve Corpus). The utterance vector $\mathbf{x}$ and the similarity vector $\mathbf{s}$ are then concatenated and passed through a single-layer MLP to predict a continuous proficiency score:
\begin{align}
\hat{y}^{\text{w2v}}_p = \text{SpeechGrader}_p(R_{A_p}).
\end{align}
\subsection{Multimodal grader}
Our Multimodal Large Language Model (MLLM) grader is built upon \textit{Phi-4-Multimodal-Instruct}, a Transformer-based neural architecture\cite{Vaswani2017Attention} adept at processing and integrating information from multiple modalities. The core \textit{Phi-4-Mini} model is a 32-layer, 3.8 billion parameter decoder-only Transformer featuring Group-Query Attention\cite{GQA} and LongRoPE positional encoding\cite{LongRope}, supporting context lengths up to 128K tokens. For speech input, a modality router and a pre-trained audio encoder project acoustic representations into the shared text embedding space. The language backbone is fine-tuned using Low-Rank Adaptation (LoRA) (rank 320)\cite{hu2022lora}, and a dedicated audio adapter (approx. 0.92 billion parameters) enhances modality-specific learning. While \textit{Phi-4-Multimodal} supports vision, we in this work utilize only its speech and text capabilities. The overall MLLM grader architecture is depicted in Figure~\ref{fig:mllm_grader}.

\begin{figure}[!t]
    \centering
    \subfloat[]{
        \includegraphics[width=0.46\linewidth]{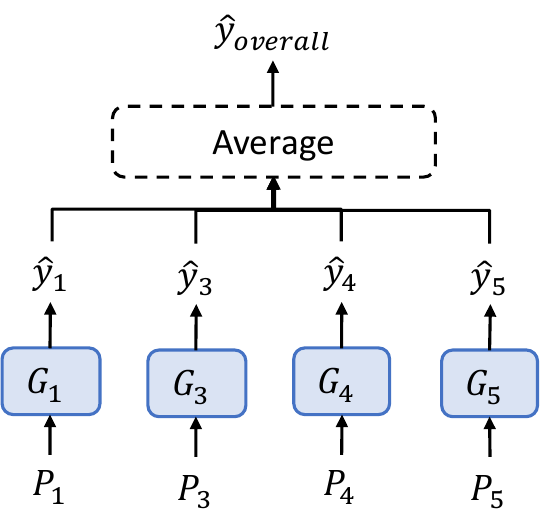}%
    \label{fig:ts_grader}}
    \hfil
    \subfloat[]{
        \includegraphics[width=0.46\linewidth]{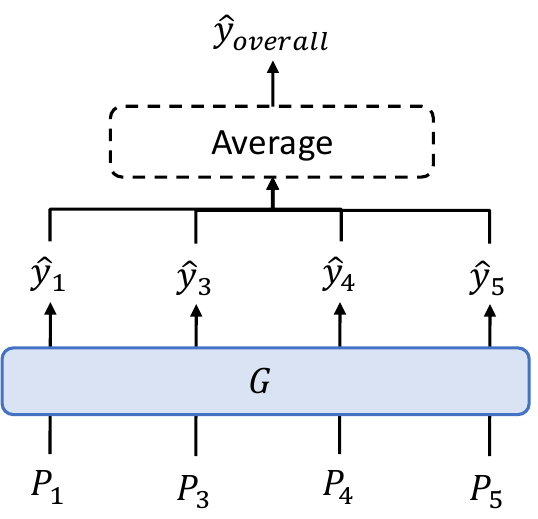}%
    \label{fig:ctg_grader}}
    \caption{Conceptual overview of two grader architectures:\ref{fig:ts_grader} Task-specific Graders, where a dedicated model ($G_p$) is trained for each task part ($P_p$), and \ref{fig:ctg_grader} CTG-MG model, where a single model ($G$) processes all task parts. In both scenarios, individual part scores ($\hat{y}_p$) are averaged to produce $\hat{y}_{overall}$.}
\label{fig:grader_architectures}
\end{figure}

For a given assessment part $p$, the MLLM grader takes various inputs—raw speech signal ($R_{A_p}$), ASR transcript ($R_{T_p}$), question prompt ($Q_p$), and task description ($D_p$)—to produce a holistic proficiency score:
\begin{align}
\hat{y}^{mllm}_p = \text{MllmGrader}_p(R_{A_p}, R_{T_p}, Q_p, D_p).
\label{eq:mllm_score_inputs}
\end{align}
Here, $R_{A_p}$ denotes raw audio and $R_{T_p}$ represents the corresponding ASR transcript. The comprehensive textual and contextual inputs are $I_p = \{R_{T_p}, Q_p, D_p\}$. Although MLLMs such as Phi-4-multimodal provide a strong foundation for task-agnostic grading because of their joint multimodal reasoning capability, we thus explore a Cross-Task-Generalized Multimodal Grader (CTG-MG). CTG-MG adopts the Phi-4 architecture as its backbone; the resulting variant is termed Phi4-Cross-Task Grader (Phi4-CTG) throughout this paper (see Figure~\ref{fig:ctg_grader}). The model was initially trained on combined data from all task parts using ASR transcripts. However, our primary strategy hones in building task-specific models to better capture the nuances of each assessment part. As such, we develop four distinct task-specific MLLM graders (dubbed Phi4-Single-Task Grader (Phi4-STG) in Table~\ref{tab:baseline_results}) based on Phi-4-multimodal respectively for each part. These Phi4-STG models prioritize the direct processing of raw audio waveforms ($R_{A_p}$), simultaneously supplemented by other input resources like question text and more. Notably, for Parts 1 and 5 of the contest, speaker context is employed by processing all of a speaker's audio samples collectively for a holistic assessment. This direct audio processing allows the MLLM to leverage its sophisticated understanding of speech nuances (e.g., pronunciation, fluency, and prosody) alongside linguistic content from its language modeling core, thereby mitigating ASR error propagation that normally occurs in cascaded systems. For any given part, at least one of $R_{A_p}$ or $R_{T_p}$ must be provided.

\subsection{Score fusion}
To integrate the complementary strengths of the speech and MLLM graders, we adopt a score fusion strategy. Specifically, the proficiency score predicted by a multimodal grader for each spoken response is first discretized into one of eight CEFR-aligned intervals: [0.0–2.25), [2.25–2.75), [2.75–3.25), [3.25–3.75), [3.75–4.25), [4.25–4.75), [4.75–5.25), and [5.25–6.0]. These intervals are derived from the annotation range used in the Speak \& Improve Corpus, which spans CEFR levels from A2 to C1+ in 0.5-point increments. For each interval, the optimal interpolation weight $w_k$ is selected via grid search on the development set, with the objective of minimizing the root mean square error (RMSE) between the fused prediction and the reference score. The weights were then fixed for evaluation. The mathematical formulation for this fusion is 
\begin{align}
\hat{y} = (1-w_{k}) \hat{y}^{\text{w2v}} + w_{k} \hat{y}^{\text{mllm}}.
\label{eq:score-fusion}
\end{align}

\section{Experiments}\label{sec:experiment}

\subsection{Dataset and evaluation metrics}\label{sec:data}

The Speak \& Improve (S\&I) Corpus 2025 \cite{Qian25-SIChallenge} \cite{ sicorpus25} contains 315 hours of open-ended spoken responses from L2 English learners with CEFR levels ranging from A2 to C1+. This corpus includes four distinct task types: Interview (Part 1), Opinion (Part 3), Presentation (Part 4), and Communication Activity (Part 5). Each response is scored on CEFR-aligned scales varying from 2.0 (A2) to 5.5 (C1+) in 0.5-point increments. The overall proficiency score is computed as the average of the four part scores:
\begin{align}
\hat{y}_{\text{overall}} = \frac{1}{4} \sum_{p \in \{1,3,4,5\}} \hat{y}_p.
\label{eq:overall_score}
\end{align}
We use the official data splits as provided in the S\&I Challenge benchmark. All evaluations follow a series of metrics: root mean-square error (RMSE), Pearson correlation coefficient (PCC), Spearman's correlation coefficient (SRC), and within $\pm0.5$ and $\pm1.0$ accuracy, with RMSE serving as the primary ranking criterion. The corpus statistics, annotation scheme, official data splits, and baseline systems are detailed in the challenge overview paper\cite{Qian25-SIChallenge}.

\subsection{Experiment setup}\label{sec:exp_setup}

We begin by describing the implementation details of all evaluated models, including the official BERT-based baseline, the wav2vec 2.0-based speech grader, and the proposed multimodal large language model (MLLM)-based graders.

\textbf{Baseline}: The setup for the official BERT-based baseline followed the description in the challenge overview paper\cite{Qian25-SIChallenge}.

\textbf{Wav2vec 2.0-based speech grader}: Wav2vec2-based models\footnote{https://huggingface.co/facebook/wav2vec2-base} were trained for 30 epochs using the AdamW optimizer (learning rate: $10^{-4}$) with a 600-step warm-up. The best-performing model was selected based on the highest macro F1 score on the development set. 

\textbf{MLLM-based graders}: The MLLMs were implemented using Hugging Face Transformers\footnote{https://huggingface.co/microsoft/Phi-4-multimodal-instruct} and the Parameter-Efficient Fine-Tuning (PEFT) library, employing bfloat16 precision and FlashAttention-2\cite{dao2024flashattention} to optimize training speed and memory usage.

For the task-specific MLLM models, fine-tuning was conducted exclusively through LoRA (rank 320, $\alpha = 640$). Optimization utilized AdamW (learning rate $5 \times 10^{-5}$), with models trained for 3–5 epochs.

For the Phi4-CTG models, which involved a single Phi-4 model fine-tuned on combined data from all parts, specific training configurations were applied. To ensure balanced representation, data was down-sampled to 150 instances per CEFR level (0.5 granularity), and audio inputs were standardized to a maximum duration of 30 seconds. These models were trained for 3 epochs using the AdamW optimizer with a learning rate of $4 \times 10^{-5}$, an effective batch size of 32, and a linear learning rate scheduler with a 10\% warm-up phase. In these Phi4-CTG experiments, we explored various input modality configurations: ASR transcription only (Phi4-CTG-TO), raw audio only (Phi4-CTG-AO), and both transcription and audio (Phi4-CTG-TA, which represents the primary configuration for our generalized model in Table~\ref{tab:baseline_results}). The performance of these configurations is shown in Table~\ref{tab:baseline_results}, with further discussion in Section~\ref{sec:results}.

\begin{figure}[!t]
  \centering
  \includegraphics[width=\linewidth]{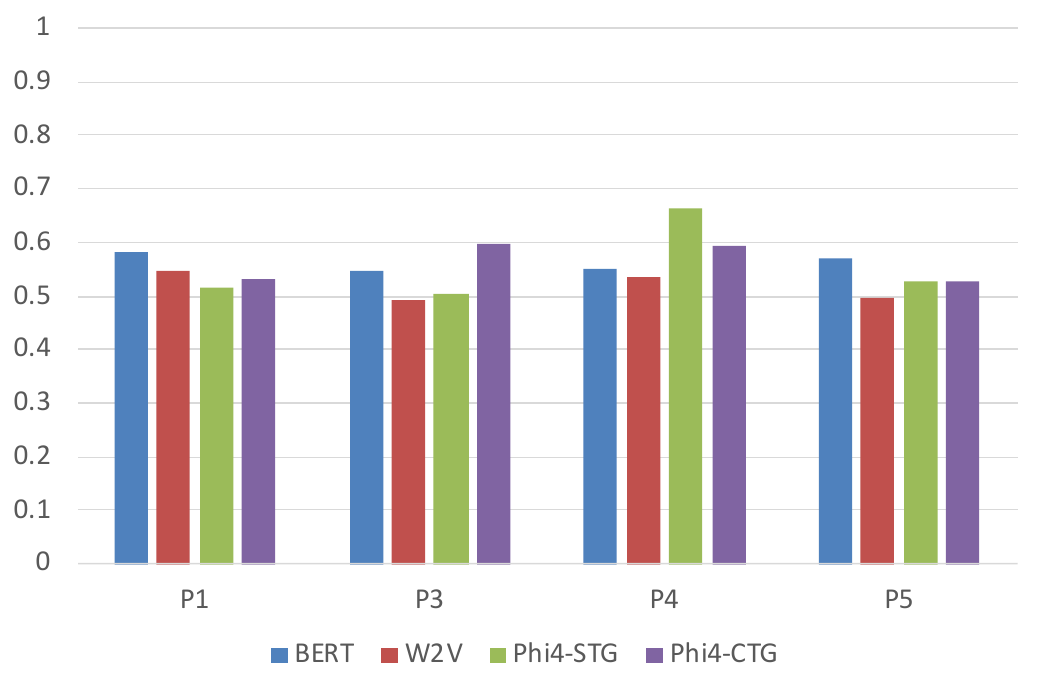}
  \caption{Comparison of RMSE across individual evaluation set components for the BERT, W2V, Phi4-STG, and Phi4-CTG.}
  \label{fig:model_cmp}
\end{figure}

\begin{table}[!t]
\caption{Experiment results on the challenge evaluation set. Phi4-STG refers to the task-specific MLLM. W2V+Phi4-STG is our final proposed integrated system.* indicate our replicate result of baseline}
\label{tab:baseline_results}
\resizebox{\columnwidth}{!}{%
\begin{tabular}{lccccc}
\hline
\textbf{Model}   & \textbf{RMSE \bad}  & \textbf{PCC \good}   & \textbf{SRC \good}   & \textbf{\%$\leq$0.5 \good} & \textbf{\%$\leq$1.0 \good} \\ \hline
BERT*             & 0.445 & 0.727 & 0.729 & 76.0 & 96.3          \\
W2V          & 0.394 & 0.790 & 0.797 & 81.3 & 99.3          \\ \hline
Phi4-CTG-TA           & 0.412 & 0.796 & 0.797 & 74.7 & 98.0          \\
Phi4-CTG-AO        & 0.428 & 0.793 & 0.795 & 75.7 & 98.0          \\
Phi4-CTG-TO       & 0.463 & 0.759 & 0.756 & 71.0 & 97.3          \\ \hline
Phi4-STG           & 0.389 & 0.808 & 0.810 & 78.3 & 98.0          \\ \hline
W2V + Phi4-CTG & 0.377 & 0.809 & 0.817 & 80.3 & \textbf{99.7} \\
W2V + Phi4-STG & \textbf{0.375} & \textbf{0.820} & \textbf{0.827} & \textbf{82.7}                        & 99.3                                 \\ \hline
\end{tabular}%
}
\end{table}

\begin{table}[t!]
  \centering
  \caption{Performance comparison with selected entries from the Speak \& Improve Challenge 2025 leaderboard (SLA Open Results, \textbf{evaluation set}).}
  \label{tab:sla_leaderboard}
  \resizebox{\columnwidth}{!}{%
  \begin{tabular}{lccccc}
    \toprule
    \textbf{Participant (Rank)} & \textbf{RMSE \bad} & \textbf{PCC \good} & \textbf{SRC \good} & \textbf{\%$\leq$0.5 \good} & \textbf{\%$\leq$1.0 \good} \\
    \midrule
    perezoso (1) & \textbf{0.364} & \textbf{0.826} & \textbf{0.830} & \textbf{83.0} & \textbf{99.7} \\
    NTNU SMIL V (2) & 0.375 & 0.820 & 0.827 & 82.7 & 99.3 \\
    nhanphan (3) & 0.384 & 0.801 & 0.815 & 81.3 & 99.3 \\
    NTNU SMIL X (4) & 0.389 & 0.808 & 0.810 & 78.3 & 98.0 \\
    Baseline (25) & 0.440 & 0.726 & 0.727 & 73.7 & 96.7 \\
    \bottomrule
  \end{tabular}%
}
\end{table}

\section{Results \& Discussion}\label{sec:results}

The experimental results are presented in Table~\ref{tab:baseline_results}, which provide a comprehensive evaluation of multiple system configurations for spoken language assessment (SLA). The official BERT-based baseline achieves an RMSE of 0.445, serving as a reference point for subsequent comparisons. The standalone wav2vec 2.0 (W2V) model outperform this baseline with an RMSE of 0.394, highlighting the effectiveness of acoustic feature modeling for SLA tasks. The task-specific Phi4-STG model, which incorporates both acoustic and semantic inputs, further improves performance, attaining an RMSE of 0.389 and a Pearson correlation coefficient (PCC) of 0.808. While W2V surpasses Phi4-STG in prediction accuracy within the $\pm0.5$ and $\pm1.0$ CEFR error margins likely due to its dedicated focus on acoustic information—Phi4-STG demonstrates stronger overall score correlations and more consistent performance across tasks.
Experiments with the Phi4-CTG variants reveal that general-purpose multimodal large language models (MLLMs) are promising but currently less effective than task-specific architectures. The Phi4-CTG-TA model yielded an RMSE of 0.412, outperforming the audio-only variant (Phi4-CTG-AO) and the transcription-only version (Phi4-CTG-TO). These comparisons reaffirm the critical importance of acoustic input for accurate SLA modeling. Among all single-model configurations, task-specific Phi4-STG emerges as the top-performing model. The consistently strong RMSE of Phi4-STG across different parts of the evaluation set (as illustrated in Figure \ref{fig:model_cmp}) further validates its robustness and motivates its use in system fusion.

The final integrated system, which applied a score-conditioned fusion of W2V and Phi4-STG (denoted as W2V + Phi4-STG), achieves the best overall performance, with an RMSE of 0.375, PCC of 0.820, and Spearman's rank correlation coefficient (SRC) of 0.827. As shown in Table \ref{tab:sla_leaderboard}, this system, submitted as NTNU SMIL V, came out in second place on the official leaderboard of the Speak \& Improve Challenge 2025. Notably, it outperforms both its individual components and alternative fusion approaches, such as W2V + Phi4-CTG (RMSE 0.377). These results substantiate the value of combining the fine-grained acoustic modeling of W2V with the contextual reasoning capabilities of task-specific Phi4-STG, resulting in a more accurate and reliable SLA system.

\section{Conclusion and Future Work}\label{sec:conclusion}

This work has presented a multimodal large language model (MLLM) system tailored for spoken language assessment on L2 learners, developed in the context of the Speak \& Improve Challenge 2025. By integrating a wav2vec 2.0-based speech grader with task-specific MLLMs through a score-level fusion mechanism, the proposed system achieved robust performance, ranking second on the official leaderboard. Experimental results have also confirmed the effectiveness of combining fine-grained acoustic representations with contextual and semantic modeling tailored to the assessment task. Comparative analysis also highlighted the superior performance of task-specific MLLMs, such as Phi4-STG, over more generalized models like Phi4-CTG, particularly in diverse SLA contexts. Moreover, the consistently strong results from models that incorporate speech input emphasize the foundational role of acoustic information in this domain. Future directions include enhancing MLLM architecture design, refining fusion strategies, and addressing concerns related to model efficiency and equitable performance across heterogeneous learner groups.

\bibliographystyle{IEEEtran}
\bibliography{mybib}
\end{document}